\documentclass[sn-aps]{sn-jnl}

\jyear{2022}%

\theoremstyle{thmstyleone}%
\usepackage{booktabs} 
\usepackage{multirow}

\theoremstyle{thmstyletwo}%

\theoremstyle{thmstylethree}%

\begin{document}

\title[Safety of autonomous vehicles]{Safety of autonomous vehicles: A survey on Model-based vs. AI-based approaches}


\author[1]{\fnm{Dimia} \sur{Iberraken}}\email{dimia.iberraken@uca.fr}

\author[2]{\fnm{Lounis} \sur{Adouane}}\email{lounis.adouane@hds.utc.fr}

\affil[1]{Universit\'e Clermont Auvergne, CNRS, SIGMA Clermont, Institut Pascal, F-63000 Clermont-Ferrand, France}

\affil[2]{Universit\'e de Technologie de Compiègne, CNRS, Heudiasyc, UMR 7253, 60203, Compiègne, France}


\abstract{
The growing advancements in Autonomous Vehicles (AVs) have emphasized the critical need to prioritize the absolute safety of AV maneuvers, especially in dynamic and unpredictable environments or situations. This objective becomes even more challenging due to the uniqueness of every traffic situation/condition. To cope with all these very constrained and complex configurations, AVs must have appropriate control architectures with reliable and real-time Risk Assessment and Management Strategies (RAMS). These targeted RAMS must lead to reduce drastically the navigation risks. 
However, the lack of safety guarantees proves, which is one of the key challenges to be addressed, limit drastically the ambition to introduce more broadly AVs on our roads and restrict the use of AVs to very limited use cases.
Therefore, the focus and the ambition of this paper is to survey research on autonomous vehicles while focusing on the important topic of safety guarantee of AVs. 
For this purpose, it is proposed to review research on relevant methods and concepts defining an overall control architecture for AVs, with an emphasis on the safety assessment and decision-making systems composing these architectures. Moreover, it is intended through this reviewing process to highlight researches that use either model-based methods or AI-based approaches. This is performed while emphasizing the strengths and weaknesses of each methodology and investigating the research that proposes a comprehensive multi-modal design that combines model-based and AI approaches. 
This paper ends with discussions on the methods used to guarantee the safety of AVs namely: safety verification techniques and the standardization/generalization of safety frameworks. }

\keywords{Autonomous Vehicles, Control Architectures, Risk Assessment, Decision-Making, Safety Guarantee, Model-based, Artificial Intelligence.}



\maketitle

\section{Introduction} \label{sec:introduction}
Transportation systems have provided the humanity priceless social and financial advantages, they are also likewise connected with negative outcomes like traffic fatalities, gas emission, and traffic congestion. Today, like never before tech companies and research laboratories throughout the world invest huge efforts to reduce these effects by automating the transportation systems. This technology has the ability to radically transform the transport sector and make roads much safer.  
Over the past decades, the increase in the vehicles number on the road led to a sharp rise in the accidents number and the automotive industry has set itself the mission of reducing this number. Advanced Driver Assistance
Systems (ADAS) came to help achieving this goal.
Nowadays, vehicles are equipped with ADAS from Level 1 and 2 according to the SAE standard grading for vehicle automation~\cite{sae2014taxonomy}. They greatly increase the safety of a vehicle and they consist in the first step toward the automation of driving functionalities to lead progressively to a fully autonomous vehicle.
According to the scientific literature, the first automated vehicle was built in Japan in 1977, within the framework of the CACS (Comprehensive Automobile Traffic Control System) project. Under the supervision of Professor S. Tsugawa~\cite{tsugawa1994vision}, demonstrations were carried out with a vehicle capable navigating in lane on its own while using a camera that detects lane markings. The vehicle was successfully driven under various road environments at the speed within 30 Km/h. 
The Defense Advanced Research Projects Agency (DARPA) Grand Challenges gave a new impulse to the research in Autonomous Vehicles (AVs)  and on the design of complex system architecture to autonomous driving. The major challenge in comparison to previous demonstrations is that there was no human intervention during all the race. Whereas the 2004 and 2005 DARPA Grand Challenges were intended to demonstrate that AVs can travel significant distances, the 2007 DARPA Urban Challenge (DUC) was designed to promote and encourage innovations in AVs in cluttered urban environments. The learned lessons from the Urban Challenge were very valuable and a lot of them are still subject of nowadays research.
Among these subjects, the guarantee of safety of AVs is one of the major research topics in the domain. It consists in providing a fully generic solution
that deals with all kinds of scenarios and is able to cope with any environment traffic condition while making the appropriate decision even in highly dynamic and uncertain environments/situations. 

\subsection{Motivations and contributions}
Driving is a complex task gathering strategic decision-making, maneuver handling and controlling of the vehicle while accounting for external factors, traffic rules and hazard. 
The purpose of researchers in this field is to develop the necessary autonomous system able to: Assess the risk in the surrounding environment; Take appropriate decision in nominal driving situation; Execute the decided maneuver; Verify the safety and coherence of the executed maneuver and Plan evasive maneuvers if required.\\
The goal through this survey is to lead the reader through the process, concepts, and methods defining an overall control architecture for AVs able to ensure high level of safety and are equipped with of an SAE level 3 or higher of autonomy~\cite{sae2014taxonomy}. We also review research on relevant methods for the safety assessment, decision-making, and guarantee of safety of AVs and propose a classification of these methods from the model-based an AI perspectives.
Indeed, one of the originalities of the paper is to highlight researches that use either model-based methods or AI-based approaches. This is performed while emphasizing the strength and weakness of each methodology and investigating the research that propose a comprehensive multi-modal design that combines model-based and AI approaches, and in certain cases both in the same structure.
After given in subsection~\ref{sec:definition}, the main useful definitions to model-based and AI-based methods, the remainder of this paper is composed of the following sections: Section~\ref{sec:control_archi} presents an overview of the most-used system architectures, in the literature, for self-driving cars. Section~\ref{sec:saf_safver} presents the related work on risk estimation and safety assessment. Section~\ref{sec:dec_mak} details research on relevant techniques for decision-making and ends with a discussion about the methods used to guarantee safety of AVs in all conditions namely: safety verification/validation and standardization/generalization of decision-making framework. A classification is proposed for each of theses parts of the different existing approaches from the model-based and AI perspectives with an emphasis on methods that have shaped the history of AVs. Finally, Section~\ref{sec:conclusion} concludes the paper.

\subsection{Main definitions} \label{sec:definition}
Model-based denotes the use of mathematical representation in the modeling of the system and thus incorporates a physical understanding of the system. Based on this understanding, vehicle’s motions and the uncertainty evolution is formalized analytically in a model-based control architecture. Among its main field, extensively used in the literature, we can cite: system identification, adaptive control, robust control, optimal control, variable structure control, Lyapunov-based controller designs, and has led to the development of frameworks such as the Robot Operating System (ROS). These methods often use statistical estimation  techniques e.g., Kalman Filtering or Particle Filtering to estimate uncertainty. 

Data-driven in the other hand is preferred when the system model is not available or hard to model, but instead the system data and historical properties are available. In fact, nowadays, a huge amount of process data is stored at every time instant. These data contain all the important state information. Using these data, online and offline, to design controller, assess risk, predict trajectories or make decision become very relevant especially under the lack of accurate models. Learning approaches, specifically deep learning, are illustrative of this approach. These examples are a part of a bigger family: Artificial intelligence (AI).

AI according to Bellman \cite{russell2016artificial} is \textit{“The automation of activities that we associate with human thinking, activities such as decision-making, problem-solving, learning, etc.”}. AI encompasses a huge variety of sub-field among which we can cite: Natural Language Processing (NLP), Knowledge representation (e.g. ontological engineering), Probabilistic reasoning and expert systems (e.g. Bayesian Networks and its variants or Markovian Processes), Learning (e.g. Supervised, Unsupervised or Reinforcement).
In this paper, our reviewing process will focus on the model-based methods from the literature compared to the AI-based methods focusing on research in learning and probabilistic reasoning for AVs.

\section{Control architectures for AVs}\label{sec:control_archi}
\begin{figure*}[h!]
  \centering
  \includegraphics[clip,width=\linewidth]{ 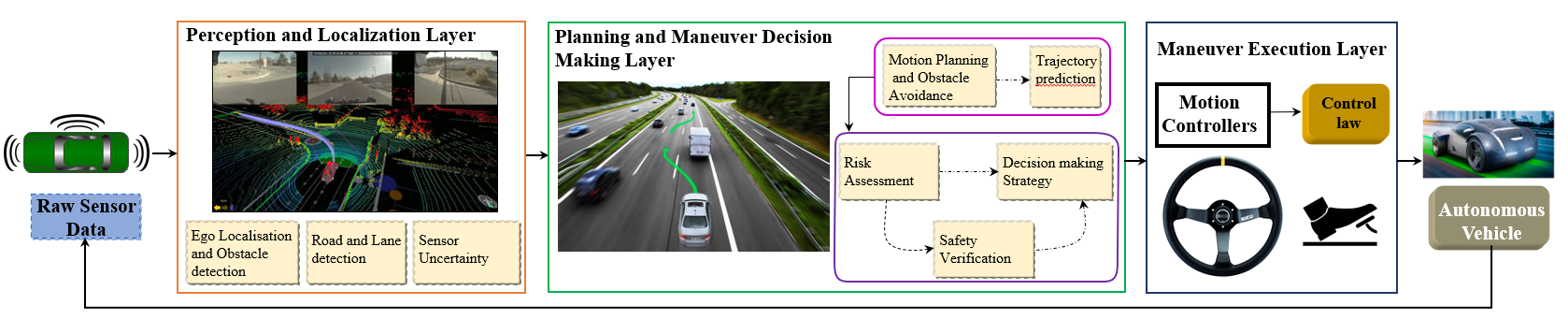}
  \caption{Standard components in autonomous driving systems listing the various tasks.} 
  \label{fig:endtoend}
\end{figure*}

Several classifications have been proposed in the literature to categorize control architectures either using model-based~\cite{BENLAKHAL2019338, pek2018computationally,shalev2017formal} or AI-based~\cite{wang2018reinforcement, hoel2018automated,Toromanoff2018} approach. Some other works try to make a classification based on the way these architectures are organized. This classification falls into two categories: modular or end-to-end control architectures.
The modular control architecture design is the most used in the autonomous driving industry. It organizes and partitions the problem of automating the driving task into a multitude of parts: localization, perception, motion planning, decision-making (also called the behavior generation/behavioral layer) and control (cf. Fig.~\ref{fig:endtoend}).
Each of these parts is divided in a multitude of sub-tasks. For example, path planning can be partitioned into trajectory prediction, obstacle avoidance, path following, behavior generation, etc. All of these behaviors are organized in a hierarchical structure able to handle the coordination while guaranteeing the stability of these systems. on the other hand, instead of keeping all the modules composing the automotive system architecture separate, an alternative framework proposes what is called ``end-to-end'' method, which integrates into one block: perception, localization, planning, and decision-making, that generates a control input for the vehicle. Typically such methods mostly rely on machine learning. 

In this section, we propose a state-of-the-art of the different architectures shown in the literature and the interactions between the modules especially motion planning, risk assessment, and decision-making, while highlighting their importance in ensuring safety. 

\subsection{Model-based system architecture} \label{sec:classical}
Many types of model-based architecture have been proposed throughout the history of mobile robotics and multi-robot systems. Centralized, decentralized, cognitive, reactive or behavioral/multi-controller architectures are examples of these architectures and have been the subject of many research in the literature~\cite{adouane2016autonomous} and have evolved with technological progress. While the reactive architectures cannot support the expanding complexity of the multitude of tasks related to autonomous navigation, the behavioral one has the ability to accomplish more complicated behaviors thanks to hierarchical coordination that selects between several elementary controllers like ADAS (such as Lane Keep Assist or Automatic Lane Change), in the purpose of mastering the overall AV behavior.
Such a behavior-based architecture and action-selection mechanism have been extensively used notably by the team VictorTango which arrived third in the DARPA Urban Challenge 2007~\cite{buehler2009darpa}. The chosen Action Selection Mechanism operates within the Behavior Integrator that chooses the winning driver among a list of drivers depending on the current situation (e.g., Merge driver, Left turn driver, etc.).\\
Optimization-based control has also been used in this kind of architecture. A pioneering event in automotive history is the Bertha-Benz historic route. In August 1888, Bertha drove her husband's vehicle the \textit{Benz Patentmotorwagen Number 3} without him knowing from Mannheim to Pforzheim, Germany over 100 km. The popular responses on her journey paved the way for the economic success of her husband and the adoption of the automobile in society.
Exactly 125 years later, a collaboration of Daimler AG and KIT (Karlsruher Institut für Technologie) automated a Mercedes-Benz S-Class named ``Bertha'' that repeated the Bertha Benz Memorial Route but this time in a complete autonomy~\cite{ziegler2014making}. The Bertha-Benz historic route is particularly challenging as it covers rural roads, 23 small villages, and major cities and passes a different variety of traffic scenarios, narrow streets, intersections and roundabouts with oncoming traffics. The autonomous vehicle had to react on a variety of objects: parked cars, bicycles, and pedestrians. 

The architecture follows the classical structure by layer~\cite{ziegler2014making}. The perception and localization modules come first and feed the lower layers with the processed perceptive information then an optimal trajectory generation is performed based on a continuous optimization. The trajectory is then transformed into actuator commands by lateral and longitudinal controllers. 
At the end of this journey, the team working on the project stated that the overall car behavior is still far inferior then the performance of an attentive human and one way to achieve comparable behavior is to improve the ability of the vehicle to interpret a given traffic scenario and predict the behavior of other traffic participants.\\
The architectures highlighted above are usually defined while using model-driven approaches to characterize the motion of the navigation system and/or the optimization functions and constraints. 
This arise the priority of having accurate models that captures and formalize the system behavior and the uncertainty evolution as modeling errors, simplifications, and linearization are among the main reasons that complicate the validation phase of AVs. This is without mentioning the impossibility and the irrelevance in certain cases of modeling large-scale systems. On this regard, the Intelligent Transportation System (ITS) community is divided as the usage of data-driven/learning approaches is seen as a cut-off with modeling. 
Even though classical architecture designs are the most used in the autonomous driving industry, the total scene understanding that is required by these architectures to derive a decision may add unnecessary complexity to the overall system when in most situations only a small portion of the detected objects are indeed relevant. In addition, the individual sub-tasks involved in each of the modules of these architectures are themselves subject to open research.\\
As mentioned earlier, the DARPA  Grand  Challenges were pioneering events in the AV field. The racing team from Stanford University won the second edition of the DARPA Grand Challenge with the robot Stanley. The winner of the race had to complete the course in less than 10 hours. The Stanford racing team had the best time with 6 hours and 53 minutes. Their main challenge was in the perception systems for road finding and obstacle detection, as well as high-speed obstacle avoidance. 
From a general point of view, Stanley’s software reflects a common approach in AV architecture design. Nonetheless, many of the individual modules relied on state-of-the-art of artificial intelligence techniques. The use of machine learning, both ahead and during the race, made Stanley robust and precise. This was a defining moment in self-driving car development, recognizing  Machine Learning and AI as central components of autonomous driving. The defining moment is additionally eminent since most of the literature work in this domain is dated after 2005 and will be discussed in section~\ref{sec:ia}.
Nowadays, a great number of control architectures have at least one module based on AI formalism~\cite{iberraken2018safe,kuutti2020survey,chen2015deepdriving}.

\subsection{AI-based system architecture} \label{sec:ia}
A certain consensus has been established in the ITS community for autonomous driving systems concerning the categorization of these systems and three paradigms emerge: The first one is the \textit{Mediated perception approaches} which analyzes the entire scene before making a driving decision. It has been discussed in section~\ref{sec:classical}. The second is called the \textit{Behavior reflex approaches} first seen in~\cite{pomerleau1989alvinn} that maps directly an input image for example to a driving action like steering and will be detailed in section~\ref{sec:endtoend}). The third paradigm proposed in~\cite{chen2015deepdriving} is called \textit{direct perception} approach and falls in between the two previous paradigms and allows the right level of abstraction. This is discussed in section \ref{sec:learning_archi} and \ref{sec:prob_archi}.

\subsubsection{End-to-end autonomous driving} \label{sec:endtoend}
\begin{figure}[t!]
  \centering
        \includegraphics[clip,width=0.8\linewidth]{ 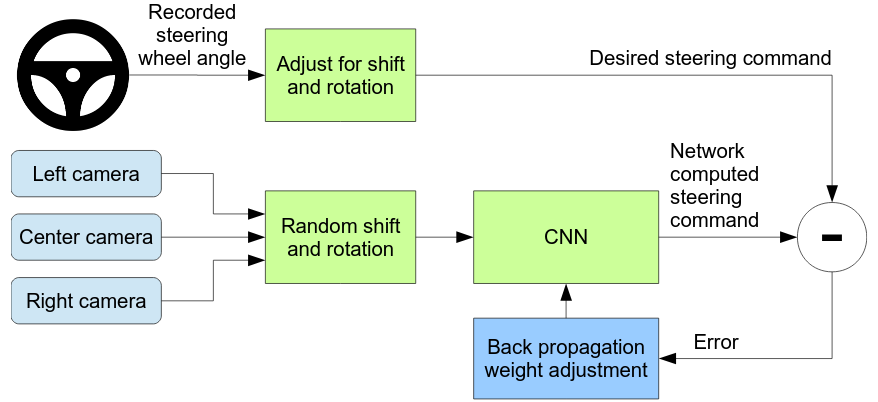}
  \caption{ System architecture of PilotNet (Image credit~\cite{bojarski2016end}). }
  \label{fig:pilotnet}
\end{figure}

Many works in the literature survey the transition between the model-based control to data-driven control~\cite{hou2013model,Di2021survey} especially with regards to deep learning applications to Autonomous Vehicle Control~\cite{kuutti2020survey}. In~\cite{kuutti2020survey}, the approaches were separated into three categories: lateral (steering), longitudinal (acceleration and braking), and simultaneous lateral and longitudinal control methods. For each of these methods, the application of deep learning has been shown.
The ALVINN (Autonomous Land Vehicle in a Neural Network) system for example proposed by Dean Pomerleau with the CMU NavLa pioneered end-to-end driving in 1989~\cite{pomerleau1989alvinn}. An artificial neural network is taught to perform lateral control by outputting steering angle to keep the vehicle on the road by taking images from a camera and a laser range finder as inputs. By 2006, it was at that point conceivable to learn how to avoid obstacles directly from raw stereo-camera inputs and that was achieved by DAVE (DARPA Autonomous VEhicle)~\cite{muller2006off}. It is trained to predict the steering angles from data provided by a human driver during training. The collected data gather a wide variety of terrains, weather condition,s and obstacle types. The learning system is a large 6-layer Convolutional Neural Network (CNN) whose inputs are unprocessed low-resolution images. The robot showed good aptitudes to detect obstacles and to navigate around them in real-time at speeds of 2 m/s.\\
Many years later, with the rise of GPU-computing capacities for efficient learning for CNN, NVIDIA ~\cite{bojarski2016end,bojarski2017explaining} made popular end-to-end methods as part of the PilotNet architecture (cf. Fig.~\ref{fig:pilotnet}). The proposed approach is to train a CNN to map raw pixels from a single front-facing camera directly to steering commands. The system learns to drive in traffic on roads with or without lane marking with only images from a front-facing camera coupled with the time-synchronized steering angle recorded from a human driver as training data. The authors stated that end-to-end system performs better than classical methods because the system's internal components are self-optimized to maximize the overall performance which is better than optimizing a selected module e.g., path planning, decision-making, etc. The motivation for PilotNet was to eliminate the process of hand-coding rules and allow the system to learn by observing.
Similar end-to-end architectures have been reported in~\cite{toromanoff2018end,chen2017end,bechtel2018deeppicar} and they differ mostly in the used sensor inputs or in the problem space.
Most of the end-to-end methods in the literature predominantly utilize Deep Neural Network (DNN) to train off-line real-world or synthetic data~\cite{bojarski2016end,toromanoff2018end,chen2017end,bechtel2018deeppicar} or Deep Reinforcement Learning (DRL) that are usually trained and tested in a simulation as the work of~\cite{jaritz2018end, perot2017end}. Some other works try to make use of DRL for real-world driving~\cite{kendall2019learning}. Pan et al.,~\cite{pan2017virtual} tried to go one step closer to real test by first training a Generative Adversarial Network (GAN) to generate real-looking images from the synthetic images of the simulation and then giving these generated images as input to the RL algorithm. Techniques for porting trained DRL models from simulation to real-world driving have been proposed in~\cite{kendall2019learning}. Imitation learning has been also widely used for end-to-end learning and is considered as the dominant paradigm in this domain. In~\cite{pan2017agile}, a model predictive controller is used as an expert to generate optimal trajectory examples and exploited to train a CNN.
Waymo through their Recurrent Neural Network (RNN) named ChauffeurNet~\cite{bansal2018chauffeurnet} proposed exposing the learner to synthesized data in the form of perturbations to the expert's driving. The authors argued that standard behavior imitation is not sufficient for handling complex scenarios. Adding these perturbations has allowed to create complex situations such as collisions and lead the learned model to be robust and able to drive a real vehicle in such critical situations.
In~\cite{codevilla2018end}, the authors state that a vehicle trained end-to-end to imitate an expert (by training a model that maps perceptual inputs to
control command) cannot be guided to take a specific turn at an intersection. They proposed instead to use command-conditional imitation learning which is an
approach to learning from expert demonstrations of low-level
controls and high-level commands input to output steering and acceleration.
In~\cite{xu2017end}, it is introduced an approach to learning a generic driving model from large scan video data set. The model used a Fully Convolutional Network - Long Short Term Memory (FCN-LSTM) architecture to learn from driving behaviors. The driving model is evaluated based on the continuous/discrete feasible action prediction across diverse conditions. In the same manner, as the previous cited research, the work given in~\cite{hecker2018end} proposed to learn a driving model using a route planner (OpenStreeMap and planned route on TomTom Go Mobile) and a surrounding view of the vehicle with a 360-degree camera input as they argue that human drivers also use rear and side views mirrors when driving.
For deeper analysis and review, the reader may refer to the extensive work and various surveys that exist in the literature concerning end-to-end learning such as~\cite{grigorescu2019survey,kiran2020deep}.
Although the idea of Dean Pomerleau~\cite{pomerleau1989alvinn} for end-to-end driving is very impressive, the need for guaranteeing functional safety is essential in self-driving cars, something that AI has difficulty with. This is mainly due to the black box problem that doesn't have the transparency of their model-based counterparts.

\subsubsection{Learning-based approaches} \label{sec:learning_archi}
\begin{figure}[th!]
  \centering
        \includegraphics[clip,width=\linewidth]{ 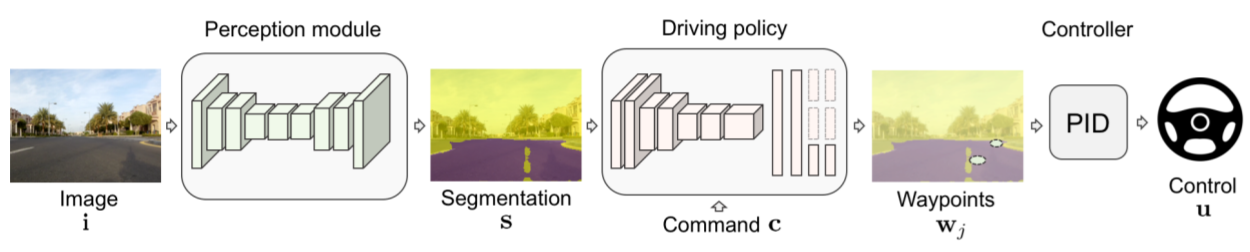}
  \caption{ System architecture by Muller et al., (Image credit~\cite{muller2018driving}). }
  \label{fig:muller}
\end{figure}

The main idea in Chen et al,.~\cite{chen2015deepdriving} is to map an input image to a small number of key perception indicators that directly refer to the accessibility of the road/traffic state for driving such as the angle of the car relative to the road or the distance to the lane marking. It is based on a deep Convolutional Neural Network (ConvNet) framework to automatically learn image features and estimate 13 indicators for driving. Based on these indicators and the speed of the car, a controller computes the driving commands to autonomously drive the car in different tracks of TORCS an open racing car simulator~\cite{wymann2000torcs}. Although the approach is very interesting, and highlights a multi-modal design that combines model-based and AI concepts, further extension is needed to handle unpredictable and complex situations.
In~\cite{muller2018driving}, it is proposed to combine the benefits of a classic driving system architecture with an end-to-end driving approach. Indeed, simulation in end-to-end driving systems brings up the problem of transferring the driving policies to the real-world. The key idea to resolve this issue is to encapsulate the driving policy such that it is not exposed directly to raw perception input. The architecture shown in Fig.~\ref{fig:muller} is organized into three major stages: perception implemented by an encoder-decoder network, a command driving policy implemented by a branched convolutional network that maps from a semantic segmentation to a local trajectory plan specified by waypoints that the car should follow, and low-level control based on a PID controller. The system is trained in simulation using CARLA simulator~\cite{Dosovitskiy17} by training a driving policy from a per-pixel semantic segmentation of the scene to output high-level control. The trained driving policy is then transferred to a robotic trick in a variety of conditions. However, the shown results were applied to a simplified scenario and further extension is needed to make it applicable to real AVs.
Some other works try to make use of what is called deep learning model (e.g., deep model predictive control~\cite{tai2016deep}~\cite{arulkumaran2017brief} or model-based Reinforcement Learning~\cite{gu2016continuous}). For instance, this method states that interactions of the system with the environment could be used to learn policies, value functions or even a model. However, learning a model introduces extra complexities and can induce model errors.


\subsubsection{Probabilistic modeling approaches}\label{sec:prob_archi}
According to Judea Pearl~\cite{pearl2018theoretical}, current AI systems only operate in a model-free mode which entails severe theoretical limits on performances as he states that such systems cannot have a retrospection reasoning and cannot thus serve a strong basis for AI. For this reason, because the purpose in the conception of an intelligent system consists in trying to imitate the inference process of humans, model-free learners need the guidance of a model of reality. He proposed thus to equip machine learning systems with causal modeling tools through graphical representation that have made model-driving reasoning computationally possible, and thus represent a good basis for strong AI.
Bayesian Networks (BNs), fall under this definition because they are considered as a probabilistic graphical language suitable for inducing models from data aiming at knowledge representations and probabilistic reasoning under uncertainty. In~\cite{lucas2001bayesian}, the authors state that BNs possess the property of being both a machine learning knowledge-based representation and a model-based formalism. Indeed, it allows structuring domain knowledge while accounting for dependencies between variables. This is also why many works classify them as knowledge-based approaches~\cite {Sankavaram2009, HAGER20101635}.
BNs have been successfully applied to solve a variety of problems in many different domains mainly related to modeling and decision-making under uncertainty~\cite{jensen2001bayesian}.  
Unlike neural networks that need extensive amounts of data and learning time, the BNs have short response time given their computational tractability (for relatively small networks) due to the exploitation of conditional independence relationships~\cite{jensen2001bayesian, schubert2012evaluating}. 
\begin{figure}[th!]
      \centering
      \includegraphics[width=0.9\linewidth]{ 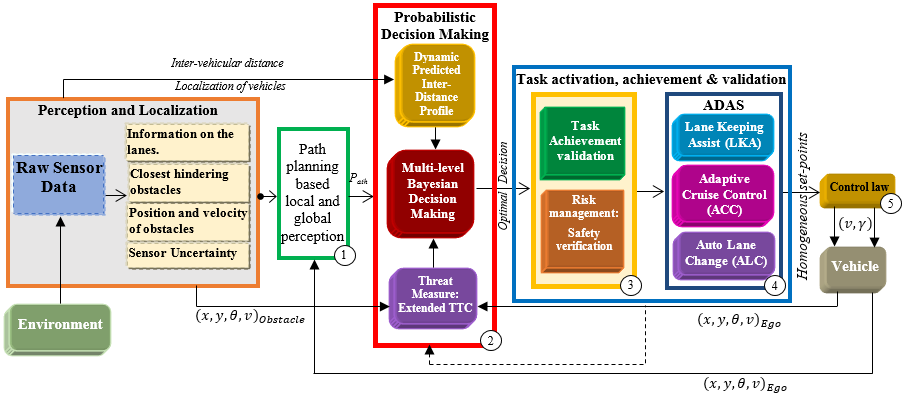}
      \caption{Probabilistic multi-controller architecture for road navigation (Image credit~\cite{iberraken2018safe})}
      \label{Archi}
\end{figure}
On the other hand, finding a realistic mathematical model that is able to understand the environment and its dynamic and make real-time decisions is not a simple task. BNs are also able to handle the uncertainty that may arise from uncertain observations or the situational model. In contrast, neural networks can solve problems with uncertainty however massive data should be available. 
In~\cite{iberraken2018safe}, it is proposed a design of a Probabilistic multi-controller architecture (P-MCA). It effectively links model-based approaches and Artificial Intelligence (AI) developments for intelligent vehicle navigation in roadway. The model-based approach appears in the path planning (based on analytical target set-points definition) and the control law (based on a Lyapunov stability analysis). The AI-based approach appears in the proposed Sequential Level Bayesian Decision Network (SLBDN) for handling lane change maneuvers in uncertain environments and changing dynamics/behaviors of the surrounding vehicles.

\subsection{Discussion}
In the light of the investigated literature, Table \ref{tab:1} illustrates a comparison between performances of the model-based and the AI-based approaches. Further, this table summarized what has been already said in each part through this chapter. Aspects considered in the depicted comparison are the most important requirements in today’s AV in terms of complexity, uncertainty handling or generalization ability. A classification of the AV problem space is proposed in connection with the mentioned requirements.\\
\begin{table*}[ht!]
\begin{center}
\resizebox{\textwidth}{!}{%
\begin{tabular}{p{2cm}|p{3cm}|p{3cm}|p{2.5cm}|p{2cm}|p{2cm}|p{3cm}}
\toprule
\textbf{Approach}       &\textbf{Algorithm Name} &\textbf{Driving environment} & \textbf{Generalization}  & \textbf{Low Computational complexity} &\textbf{Uncertainty: model errors/noisy measurements} &\textbf{Experiments} \\ \midrule


\multirow{3}{=}{\textit{\textbf{Model-based approaches}}}& Behavioral/multi-controller architectures~\cite{buehler2009darpa} & Autonomous driving in urban environment & + +  &  + +    &  -  & Real \& Simulated \\  \cline{2-7}  
                                      & Bertha-Benz Memorial Route in 2014~\cite{ziegler2014making}  & Autonomous driving in urban environment  &    +  +  &    +   &  -  &  Real \& Simulated  \\ \midrule

\multirow{5}{=}{\textit{\textbf{AI-based approaches}}} & Probabilistic Multi-Controller Architecture based on BNs~\cite{iberraken2018safe} & Autonomous driving in road-way  &    +   &    + +  &  + + & Real \& Simulated \\ \cline{2-7} 
                                   & The direct perception approach~\cite{chen2015deepdriving} & Autonomous Driving in multiple configurations   &  +  &  +    & - -  &  Testing on real-world data and on TORCS simulator~\cite{wymann2000torcs}   \\ 
                                   \cline{2-7}
                                    & PilotNet by NVIDIA (end-to-end)~\cite{bojarski2016end} & Autonomous driving in real traffic situations  &  +  &   - -   &   + + & Real \& Simulated  \\ \cline{2-7}
                                    & ChauffeurNet by Waymo (end-to-end)~\cite{bansal2018chauffeurnet} &  Autonomous Driving in multiple urban configurations     &  +      &   - -  &   + + & Real \& Simulated  \\ 
                                                 
\bottomrule

\end{tabular}}
\caption{Comparison between model-based and AI-based system architecture for autonomous vehicles.}\label{tab:1}
\end{center}
\end{table*}
Accordingly, model-based approaches have a strong ability to control vehicles when accurate models are available and they also have a systematic modular design which makes it possible to improve or add additional features not supported in the original design without huge modification or training. Generalization regardless of the used method is a challenge in the AV domain due to the highly complex nature of the navigation environment, however model-based does well compared to others as the algorithms are analytically defined. 
Learning methods have a better performance when the models are not available but they lack systematic designing procedures and means of analysis. Specifically, end-to-end approaches on the other side have the strength to be compact and self-optimized as all the modules composing the classical architecture are integrated. 
However, in real-world applications, data is limited in quantity and quality and is usually gathered for a specific task or scenario. Path planning, trajectory prediction, and control demand real-time performance, which means for deep learning usage, the time-consuming data collection and training procedure should be simplified for online systems. Even with these conditions, proof of the stability of the control system is necessary and how to generalize the deep learning models to all cases is still a challenge~\cite{tai2016deep}. Indeed, a single false decision can lead to a critical situation. 
A promising paradigm arose and aims to reach the defined objectives while using a smart combination of AI-based and model-based formalism. These approaches have the utility of dealing with the limitations that arise from each one of the methods. If a good equilibrium between approaches is found, we believe that this can be the basis of a powerful system.

\section{Risk Assessment of AVs}\label{sec:saf_safver}
While it may not be difficult for human drivers to tell whether a situation is safe, it is far from obvious for an autonomous car~\cite{guo2019safe}. In that context, it is associated with the idea of whether the situation is or will be risky/dangerous for the vehicle and for other traffic participants. Thus, it is natural to consider collisions as the main source of risk and to base the assessment of risk solely on collision prediction~\cite{lefevre2014survey}. Therefore a maneuver is said to be safe if no potential collision is possible and risk could be intuitively understood as the likelihood and severity of the damage that a vehicle of interest may suffer in the future. From this understanding, to assess risk it is necessary to predict how a particular situation will evolve. This is performed while using motion prediction~\cite{katrakazas2019new,kim2017collision,sierra2019towards}. It is used to infer the intentions of the surrounding drivers and predict what their state will be in the future timesteps. This domain has been the center of interest of numerous works in robotics~\cite{lefevre2014survey, paden2016survey,gonzalez2015review}. Lefevre et al.,~\cite{lefevre2014survey} proposed a comprehensive survey that classified existing motion prediction approaches used for risk assessment of AVs into three distinct parts: Physics-based motion model, Maneuver-based motion model, and Interaction-aware motion model. Table~\ref{tab:2} represents a summary of motion models and prediction inspired by Lefevre's classification~\cite{lefevre2014survey}. After predicting the potential future trajectories of all the moving entities the next step is to effectively detect collisions between pairs of entities and their predicted trajectories. This is known as \textit{Collision-based risk assessment}.
\begin{table*}[t!]
\begin{center}
 \resizebox{\textwidth}{!}{%
\begin{tabular}{p{4cm}|p{4cm}|p{4cm}|p{4.2cm}}
\toprule
\textbf{Approach}   &\textbf{Methods}   &\textbf{Algorithm and tools} &\textbf{Advantages (+) and limitations (-)}  \\ \midrule

\multirow{3}{=}{\textit{Physics-based motion model~\cite{lefevre2014survey,sierra2019towards}}}  & Kinematic and Dynamic models & Kalman Filtering, Monte-Carlo simulation & + Simplicity and Computational efficiency        \\ 
                                       &     &      & - Short-term prediction  \\ 
                                      &     &      & - Inter-Vehicles interaction disregarded   \\ 
                                    \midrule

\multirow{3}{=}{\textit{Maneuver-based motion model~\cite{houenou2013vehicle,sierra2019towards,zyner2018recurrent}}}   & Maneuver Intention estimation coupled with trajectory prediction   &  Most popular method is Hidden Markov Model (HMM), Classification of motion patterns~\cite{lefevre2014survey}, Support Vector Machines (SVM)~\cite{mandalia2005using}   &  + Longer prediction horizon ~~~~~~~~~~~ - Inter-Vehicles interaction disregarded \\ 
                                                
                                                \midrule
\multirow{3}{=}{\textit{Interaction-aware motion model~\cite{schulz2018interaction,ward2017probabilistic}}} & Trajectory prediction by including inter-vehicles interaction  & Majority are build using Dynamic Bayesian Networks (DBNs)~\cite{katrakazas2019new} &    + Consider interaction between road participants                                    \\ 

                                      &     &   &  + Design Flexibility
                                     \\ 
                                   &  &    &     - Computationally expensive   \\

\bottomrule

\end{tabular}}
\caption{Motion Models and Prediction: A summary (updated version of~\cite{lefevre2014survey})}\label{tab:2}
\end{center}
\end{table*}
While the simplest techniques provide basic methods on whether and when a collision will occur, more complex methods can compute in addition information on its probability or its severity.
However, these collision-based risk indicators rely on a rather unconstrained evolution of the environment. It estimates the risk of a situation by predicting the future trajectories mostly based on the current state and then looking for a collision between these trajectories. 

Another interesting approach for safety assessment named \textit{Behavior-based risk assessment}, estimates the risk of vehicles deviating from their nominal behavior expected on the road.
Hundreds of researchers have been done in the domain of warning and detection systems~\cite{4597085,sikander2018driver} of incoherent or unexpected events done by vehicles in the environments mostly for ADAS. When dealing with AVs, the behavior-based risk can mostly be estimated by either: Defining a nominal behavior of vehicles and then detect events that do not match, detecting conflicting intentions between vehicles or with regards to traffic rules, or by dealing with unexpected behavior as independent events while using safety verification techniques, upstream of the risk assessment modules for emergency situations which will be discussed in Section~\ref{sec:safety_ver}. 
Many methods have been proposed in the literature and we propose in what follows, a classification of the risk assessment methods from model-based and AI-based perspectives. 

\subsection{Model-based approaches}
In this section, the model-based techniques will be discussed. These techniques generally use the physics-based motion models of evolution (cf. Table~\ref{tab:2}) in order to assess the risk. The risk metrics are divided in two categories: the deterministic risk indicator and optimization-based risk assessment. 

\subsubsection{Deterministic risk indicators}
It relies on the \textit{collision-based risk assessment} definition. 
The most-known indicators of criticality are: the change in velocity of the vehicles, the collision angles (rear-end or head-on), the amount of overlap between different shapes representing vehicles (ellipses, circles, polygons, etc.)~\cite{adouane2016autonomous, pek2018computationally, hou2014new}, the occupancy of conflicting areas~\cite{schulz2018interaction}, the rate of change steering, the configuration of trajectories in a collision course, the remaining time span in which the driver can still avoid a collision by braking (e.g., Time-to-Brake or by steering, etc.) 
In the literature, there exists several risk metrics ~\cite{keller2011active,brannstrom2010model,berthelot2011handling,sandblom2011probabilistic,hillenbrand2006multilevel,noh2017decision} that cover these indicators of criticality. Table~\ref{tab:risk_deter} presents some of the most-known deterministic risk measures used in nowadays AVs. They are classified in a chronological way of appearance in the literature as well as in three categories which are: the longitudinal risk assessment, lateral risk assessment and both at the same time.
\begin{table*}[hb!]
\caption{Some examples of deterministic risk measures}\label{tab:risk_deter}
\begin{center}
 \resizebox{\linewidth}{!}{%
\begin{tabular}{p{3cm}|p{4cm}|p{5cm}|p{4cm}}
\toprule
\textbf{Approach}&\textbf{Risk measure}   & \textbf{Description}   & \textbf{Equation}  \\ \midrule
\multirow{15}{=}{Longitudinal Risk Assessment} & Time-to-Collision (TTC) (Hayward 1972)~\cite{hayward1972near}  & The time required for two vehicles defined with their pose, velocity, and wheelbase denoted as $(X_i, V_i, L_i)$  to collide if they continue on at their present speed and path.  & $TTC = \frac{X_1- X_2-L_1}{V_1-V_2}$ 

if $V_2 > V_1 $ for the case of rear-end collision (see~\cite{laureshyn2010evaluation} for other configurations).   \\  \cline{2-4}   
& Post-Encroachment Time (PET) (Allen et al., 1977)~\cite{allen1978analysis}  &  The time between the moment $t_1$ that the first vehicle last occupied a position and the moment $t_2$ that the second reaches the same position. &  $PET= t_2 - t_1$   \\ \cline{2-4}
& Deceleration to Safety Time (DST) (Hupfer, 1996)~\cite{hupfer1997deceleration}  &  The deceleration $a_{DST}$ that has to be applied to the \textbf{e}go vehicle velocity $v_e$ to maintain a certain safety time $t_s$ with respect to the \textbf{o}bject vehicle velocity $v_0$.  & $a_{DST} = \frac{3(v_e-v_0)^2}{2(x-v_0.t_s)}$

with $x$ the distance between both vehicles. (see~\cite{schubert2012evaluating} for more details)     \\    

\midrule

\multirow{3}{=}{Lateral Risk Assessment} & Time to Line Crossing (TLC) (Godthelp 1984~\cite{godthelp1984development})   & The time duration $t_{LC}$
available for the driver before any lane boundary crossing. &  For straight road configuration and zero steering angle:

$t_{LC}=\frac{y_{ll}}{v_l}$ with $v_l$ the lateral velocity and $y_{ll}$ the lateral distance of front left tire to the line that would be crossed.
(see~\cite{godthelp1984development,mammar2006time} for more configurations)     \\   \midrule
\multirow{10}{=}{Longitudinal \& Lateral Risk Assessment} & Time To React (TTR) (Hillenbrand et al., 2006) ~\cite{hillenbrand2006multilevel}  &  The remaining time to avoid an imminent collision by emergency braking with full deceleration (which is known as Time To Brake (TTB)), steering with maximum lateral acceleration (Time-To-Steer (TTS)), or a kick-down maneuver (Time-To-Kickdown (TTK)) by leaving the collision zone early enough for example. & The TTR is calculated as the maximum of the TTB, TTS and TTK~\cite{hillenbrand2006multilevel}      \\ 
\bottomrule
\end{tabular}}
\end{center}
\end{table*}
A unique warning threshold is usually applied to the obtained value to classify the situation as risky or safe. Some other works~\cite{lee2004comprehensive, Wang2018164} distinguish multiple warning levels that improve the decision-making process and conform to the driver perception of safety in a dynamic environment.
Despite the good number of metrics that exist in the literature, the Time-to-Collision (TTC) remains the most used and the oldest risk metric to the best our knowledge in the domain of AVs. This is mainly because of its simplicity, its low cost computational time and it has many variants for multiple different configurations~\cite{laureshyn2010evaluation}.
However, as pointed out by Laugier et al., in~\cite{laugier2011probabilistic} this metric suffers from the lack of context and singularities that may occur in some configurations. Indeed, TTC alone is insufficient as a risk indicator for managing complex situations. In addition, the common definition of the TTC is restricted for a specific path to detect longitudinal collision and for well-defined scenarios such as car following.
To overcome this issue, extended definitions of the TTC have been proposed in multiple works~\cite{minderhoud2001extended,jimenez2013improved,hou2014new,ward2015extending}. The works given in~\cite{iberraken2018safe} and~\cite{hou2014new} address the problem from a planar perspective where vehicles are considered in a two-dimensional plane and the state of each vehicle is defined by a vector position, velocity and acceleration components on X and Y direction. This Extended TTC (ETTC) is computed at each time step for each vehicle pair that are close enough.
Even though a deterministic approach has less computational complexity, the main drawback of these criteria is the incapability to deal with the unpredictable as only short-term prediction is considered by these metrics.

\subsubsection{Optimization-based risk assessment}
\begin{figure}[h!]
    \centering
    \includegraphics[width=0.7\linewidth]{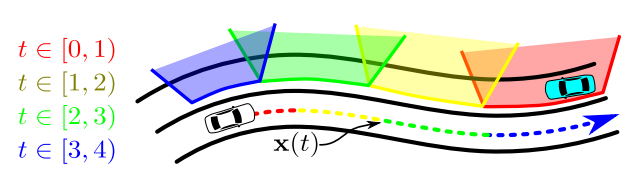}
    \caption{Constraints for an oncoming Object (cyan). The trajectory is constrained by polygons of corresponding color (Image credit:~\cite{ziegler2014making}).} \label{fig:berth_plann}
\end{figure}

Other methods estimate the risk jointly within the path planning through algorithms (for example while using optimization approaches) based on a chosen trajectory considered as safe with respect to certain constraints related to the vehicles dynamic, the road geometry, the dimension of the vehicle or the occupancy of objects in the environment.
In this kind of application one can make the analysis concerning the constraints defined in the optimization and the used algorithm.
On-board of the vehicle that completed the 103 km of the Bertha-Benz-Memorial-Route fully autonomously, Ziegler et al.,~\cite{ziegler2014trajectory} proposed to use an optimal trajectory generation based on a continuous optimization.
The solution trajectory is the constrained extremum of an objective function that is designed based on the dynamic feasibility and comfort. Static and dynamic obstacle constraints are incorporated in the optimization in a form of polygons (cf. Fig.~\ref{fig:berth_plann}). The constraints are designed in such away that the solution converges to a single global optimum. 
Pek et al., in~\cite{pek2018computationally} developed a fail-safe trajectory planner for self-driving vehicles. This trajectories are computed in real-time in continuous space by making use of convex optimization techniques. This allows to separate motions into a longitudinal and a lateral component while defining the constraints suitable for each motion and thus guaranteeing the drivability of the resulting motions. Collision avoidance is done through the convex constraint set while considering the kinematic vehicle model. It is described with respect to a curvilinear coordinate system aligned to a reference path and restricts solutions so that they do not intersect with the predicted occupancy sets of other traffic participants. 

\subsection{AI-based approaches}
In this section the AI-based techniques will be discussed and are divided in two categories: the probabilistic methods and the learning approaches. 

\subsubsection{Probabilistic methods for risk assessment}
These approaches take into account the uncertainty of motion along the predicted trajectory (cf. Table~\ref{tab:2}). It generally uses the maneuver-based or the interaction-aware motion model (cf. table~\ref{tab:2}) in order to assess the risk. Several probabilistic frameworks have been used: Hidden Markov Model (HMM)~\cite{firl2011probabilistic}, DBN~\cite{li2019dynamic} or other probabilistic frameworks.
In~\cite{katrakazas2019new}, it is proposed to integrate what is called in the paper a network-level collision prediction through a Random Forest classifier with interaction-aware motion models under a Bayesian framework DBN for risk assessment of AVs. The network-level collision prediction consist of the safety context (safe or collision-prone) of the road segment on which the ego-vehicle is traveling on and interprets the traffic scene as either ``dangerous'' or ``safe'' enabling a vehicle to be more vigilant in collision-prone situations. \\
The grid based approach is another way to assess the risk. It consists in constructing grid cell values from sensory information. It is used to model the environment and propose to split the space into a set of cells that may be free or occupied. Usual methods aim to calculate the probability of occupation of a cell from sensor data. It was first proposed by Elfes in~\cite{elfes1989using}. 
Bayesian Inference is the common used methods to cope with uncertainty and errors. Many extension have been published in the literature. For example the Bayesian Occupancy Filer (BOF) used by~\cite{coue2006bayesian} provides filtering, data fusion, and velocity estimation capabilities of the cells while allowing parallel computation.
Evidential grids~\cite{moras2011credibilist,pagac1998evidential} are a variation of occupancy grids. It is is based on Dempster-Shafer (DS) theory and offers a solution to make the difference between unknown and doubt caused by conflicting information output from the fusion process. Many other works combine the efficiency of a deterministic criteria like the TTC with an interaction-aware formalization while using a DBN. It is the case for example of the works done in~\cite{katrakazas2019new,gindele2015learning}.\\
Other researches used the reachable set computation in order to assess collision in future planned path~\cite{althoff2009model,BENLAKHAL2019338}. The classical definition of a reachable set is a set that contains all possible states that the system trajectories can evolve into. A collision is verified by checking whether the reachable positions of the ego vehicle intersect the reachable positions of another vehicle.
However, with the classical definition of reachable set, which could be considered as too conservative, other vehicles rapidly cover all positions the AV could possibly move to. Which results in the fact that the planned paths of AVs are often evaluated as unsafe. For this reason, the reachable sets are enhanced by stochastic information that gives the probability of the crash and it is called \textit{stochastic reachable sets}~\cite{althoff2008stochastic}.\\
In order to avoid the extensive complexity of generating all the possible trajectories to detect collisions between each possible pair that is often used in \textit{Collision-based risk assessment}, \textit{Behavior-based risk assessment} is used. 
For example, Lefèvre et all., in~\cite{lefevre2012risk} proposed an interaction-aware formalism for the trajectory prediction and the maneuver intention prediction. Dangerous situations are identified by detecting conflicts between intention and expectation, i.e., between what drivers intend to do and what is expected of them. It is formulated as a Bayesian inference problem where intention and expectation are estimated jointly for vehicles converging to the same intersection. Risk is computed as the probability that expectation and intention do not match. This method can be categorized in both behavior-based and probabilistic risk assessment as the risk is computed as the probability that expectation and intention do not match. 
Another example is the work of Schulz et al.,~\cite{schulz2018interaction}. 
By making use of the interaction aware formalism, it was proposed a behavior prediction framework through a DBN, which explicitly considers  the intentions of drivers and the inter-dependencies between their future trajectories. The decision-making process of an agent is divided into three hierarchical layers: which route the vehicle is going to follow (route intention), whether it is going to pass a conflict area at an intersection before or after another agent (maneuver intention), and what continuous action it is going to execute. Unexpected behaviors are not included explicitly in the framework however the inter-dependencies between all the vehicles future trajectories are taking into account which allows to deal with conflicting areas and act accordingly.\\
\subsubsection{Safety of learning approaches} 
Control architectures in AI are either solely based on machine learning in an end-to-end fashion (cf. section~\ref{sec:endtoend}) or involve some combination of model-based reasoning and AI components (cf. section~\ref{sec:ia}). Either way, the urge for explainable AI able to guarantee safety is receiving increasing attention in nowadays researches~\cite{gunning2017explainable, huang2017safety,burton2017making,varshney2016engineering, amodei2016concrete,seshia2016towards}. More and more AI and deep learning strategies are effective and reliable even for safety-critical related issues~\cite{muhammad2020deep,nascimento2019systematic}.\\
Guaranteeing the safety of a system running any kind of method or technique heavily rely on: the type of the used technique, the application context, the understanding of the impact of possible failures, the definition of what is a safe behavior, the definition of the assumptions and the constraints on the system and its environment, the uncertainty handling. 
In the particular case of deep learning approaches, in addition to the previous requirements, the need of a clear justification about the taken decision and the insurance of safety on all possible driving situations remains subject of highly active nowadays research.
Many works in the literature try to map these requirements for deep learning techniques by defining a set of constraints and assumptions in the design of the used network that needs to hold in order to ensure the safety of the specified behavior. The work given in~\cite{burton2017making} for example, designed an application of CNNs to detect (i.e., classify and localize) objects based on camera images as part of a collision avoidance system for self-driving vehicles. For example, standard specifications which could be obtained at the level of the machine learning feature could include: the class of the located object that are at a specified distance or with a certain lateral precision and that depends on the velocity of the ego vehicle or the TTC. These assumptions are mapped to the dimensions of the image frames and presented to the CNN.\\
On the other hand, some researchers focus their effort in analyzing the faults and failures caused in deep learning components. They may be due to unreliable or noisy sensor signals, neural network topology, learning algorithm, training set, or unexpected changes in the environment. For example, Varshney in~\cite{varshney2016engineering} reasoned about the definition of the term \textit{``safety''} stating that it should be defined in terms of risk, epistemic uncertainty, and the harm incurred by unwanted outcomes. Instead of reasoning about the input data, he analyzed the choice of cost function, the appropriateness of minimizing the empirical average training cost and the empirical risk and defined strategies to achieve safety.
In the same spirit, the work given in~\cite{amodei2016concrete} attempted to analyze the problem from the point of view of  \textit{``accidents''} stating that these dangerous behaviors may arise from poor AI system design and proposed a list of five practical research problems related to accident risk, classified according to whether the problem results from having the wrong objective function (avoiding side effects and avoiding reward hacking), an objective function that is too expensive to evaluate frequently (scalable supervision), or undesirable behavior during the learning process (safe exploration and distributional shift).
The most cited example is the Tesla Autopilot accident in 2018, caused by a misclassification error despite the 130 million miles of testing and evaluation under extremely rare circumstances. Despite the fact that fail-safe mechanisms (that must stop the AV in case of a failure is detected) exists~\cite{chakarov2016debugging}, one of the main cause that lead-vehicle governed by deep learning to crash are the mistakes done by lower-level components. These mistakes propagate up to the decision-making process and lead to devastating results. In the case of the Tesla, the low-level component failed to distinguish the white side of a turning trailer from a bright sky. In this kind of system, the uncertain information, in this case distinguishing between the sky and another vehicle, should be escalated to a higher level decision and may advise the user to take control of steering. Gal in his Ph.D~\cite{gal2016uncertainty} worked on developing tools to obtain uncertainty estimation and confidence models in deep learning, casting recent deep learning tools as Bayesian models without changing either the models or the optimization.
For the purpose of accounting for uncertainties,~\cite{mcallister2017concrete} proposed to use end-to-end Bayesian deep learning architecture for AVs. A comparison has been made between a framework built on traditional (non-Bayesian) and Bayesian deep learning. Although both systems use the same initial sensory information, propagating uncertainty through the prediction and decision layers allows the Bayesian approach to avert disaster. \\
In all these cases, the safety of a decision in deep learning can be reduced to ensure the correct behavior of the system, and many surveys the techniques used and the uncertainty handling~\cite{grigorescu2019survey, gal2016uncertainty}. Verifying the program is working as intended before deployment is possible, however, safety insurance and formal verification methodologies are still not clearly fixed yet.

\subsection{Discussion}
\begin{figure}[h!]
    \centering
    \includegraphics[width=0.6\linewidth]{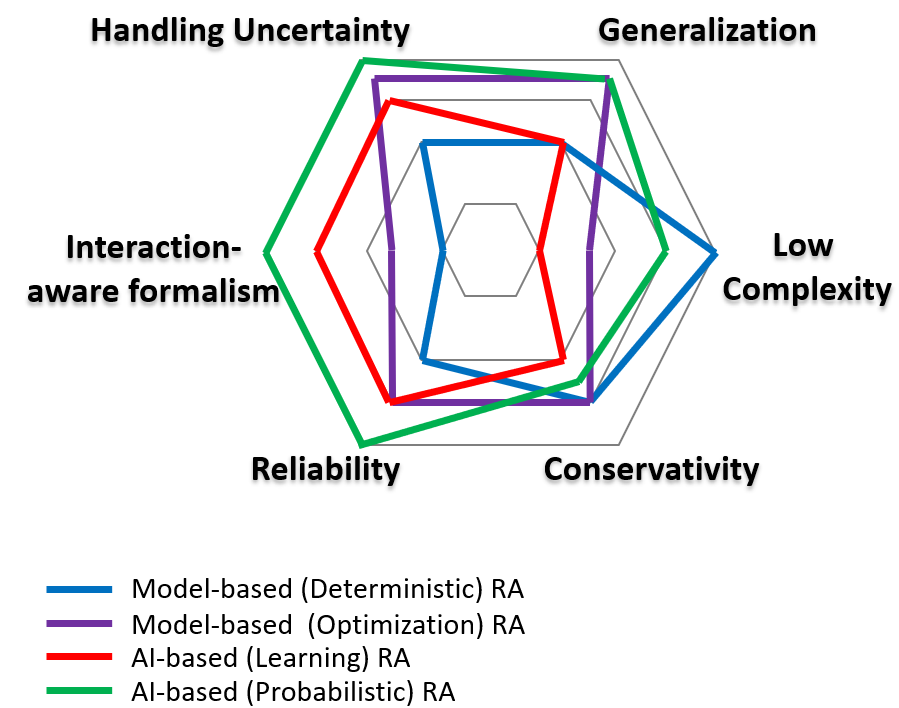}
    \caption{Comparison between the reviewed Risk Assessment (RA) methods} \label{fig:risk_compar}
\end{figure}

An important challenge in the field of risk assessment is to find the perfect balance between ensuring safety with all the imposed constraints: \textit{Uncertainties} in terms of model errors or noisy measurements, \textit{Computational complexity} assessing the risk while keeping the complexity manageable and \textit{Conservativity} in the navigation. For these reasons defining risk assessment strategy that must be able to: insure \textit{High Reliability}, take into account the prediction and the interactions between vehicles (\textit{Interaction-aware Formalism}) and \textit{Generalize} to arbitrary driving conditions and unknown environments.
Because no standardized performance indicator exists in this domain, a classification is shown in Fig.~\ref{fig:risk_compar}, based on the criteria stated above, which are considered the most used in the ITS community. 

 Obviously, none of the methods seems to be fully infallible. The AI-based (Probabilistic) seems more efficient than the remaining methods with respect to most requirements as it has the potential to consider the nature of the stochastic dynamics of the traffic environment, be able to account for model and sensor uncertainties through well-known probabilistic algorithms and consider for present and future interactions between participants. The learning methods are more and more effective and reliable even for safety-critical related issues and have the strength that they use real data in their training which makes the resulting safety system very realistic. However, this dependency on the data makes it unpredictable regarding to unknown/unseen situations for most of the defined requirements.
For the model-based RA, the deterministic ones provide low computational complexity, however, they are inefficient when dealing with the unpredictable in long-lasting maneuvers. 
The optimization-based RA is more efficient in this regard, however, these methods are dependent on the modeling of the optimization problem and its defined constraints. 

\section{Decision-making and risk management for AVs} \label{sec:dec_mak}
The need of an efficient Decision-Making System (DMS) is rising as the ultimate challenge in nowadays research. The main reason is that decision-making is located at the highest level of the automotive architecture. An efficient DMS requires well-functioning sensors and perception, self-localization, precise maps, accurate interpretation and assessment of traffic situations to make safe and efficient decision. In this section, the focus will go to risk management strategies defined within the standard modular control architecture (cf. Fig.~\ref{fig:endtoend} (a)) as end-to end methods have already been discussed in section~\ref{sec:endtoend}.

In order to provide safe and reliable AV maneuver decisions under various driving situations, four interconnected systems are needed:
Global motion planning, behavior reasoning (or risk assessment coupled with decision-making), local motion planning and control. Global motion planning finds the fastest and safest route on the road network to get from origin to the final destination. Behavior reasoning assesses the driving situation (through risk assessment cf. section \ref{sec:saf_safver}) and determines the overall behavior of the autonomous vehicle (through decision-making strategies) based on the global route and perception information. The local motion planning generates the trajectory based on the global route and the determined behavior and is responsible of avoiding static and dynamic obstacle collisions. Finally, the AV has to be controlled in order to be guided along the planned trajectory. Many works in the literature surveys motion planning algorithms for self-driving cars~\cite{gonzalez2015review,claussmann2019review,paden2016survey}, and different techniques and a number of classifications have been proposed for each algorithm. This section will focus on the decision-making strategies used for autonomous vehicles' navigation.

\subsection{From the earliest approaches to current challenges in decision-making}\label{sec:rule_dec}
Already in the early 90's, the reflection was already ongoing and different systems for decision-making of AVs were proposed. For example, one of the earliest work was the decision-making model proposed by Dickmanns et al.~\cite{dickmanns1994seeing}, in which a rule-based structure would enable alternative maneuvers for the ego-vehicle triggered by special events recognized through vision.\\
Through the different DARPA Grand Challenge (2004, 2005 and 2007) (cf. section~\ref{sec:classical}), multiple pioneering solutions have been proposed in multiple areas. But the decision-making as it is defined in nowadays research have been investigated only in the last edition. The objective was to prove that navigation could be done in traffic when there were both moving intelligent vehicles and moving vehicles driven by humans. Competitors had to drive 97 km through urban environments, interact with other traffic participants and obey to the California traffic rules. This made it necessary to the competing team to use evolved decision-making mechanism to guide the behavior of the controlled vehicles.
It is important to mention that the most common approach in the DARPA competitions was to use a rule-based decision process based on Finite State Machine (FSM) as decision-making strategy~\cite{kress2008automatically}. As an example Junior and Odin (the second and third ranked in the competition) used FSM to govern their vehicles' behavior. The FSM is given by a finite set of states in which the agent can be, and by the transitions between the states in response to some inputs. Based on the state of the vehicle in the world, the FSM searches for a suitable behavior (changing lane, making a U-turn) that makes the vehicle able to reach an objective checkpoint. Then, it either sends the controller the chosen trajectory that in return send steering and velocity commands or send a message to the planner that the checkpoint cannot be reached. Most of these transitions are coded and tested by hand and thus prone to errors. 

FSM methods perform well in basic scenarios, however, this approach lacks the ability to generalize to unknown situations and to deal with uncertainties. In addition, when considering traffic scenarios, unexpected behaviors that have not been considered during the construction of the system, which necessitates the addition of new rules, and consequently increases the complexity of the decision-making process. \\
For this reason, state machines based methods have been improved and fused with other methods to cope with a larger variety of real urban traffic scenarios~\cite{okumura2016challenges, ziegler2014making, jo2015development}.
As an example, a research using the concepts of state machine was the team working on the Bertha Benz Memorial Route~\cite{ziegler2014making}. However, they structured the program flow of upcoming behaviors/decisions while using manually implemented architecture.
The authors proposed a hierarchical concurrent state machine as behavior generation method. Depending on the current driving situation, behavior generation formulates constraints (which is considered as the program flow) that arise from the current driving corridor, static obstacles, dynamic objects, and yield and merge rules. What is called decision-making is skipped in their approach and theses constraints are directly given to the trajectory planning modules. In this case, planning is used to make decisions in the specific situation by the means of an optimal trajectory generation based on a continuous optimization.\\
In the same spirit, multiple works focus on the definition of constraints that condition the trajectory planning module rather then defining a proper decision-making module. The trajectory planner chooses a trajectory considered as safe with respect to certain constraints related to the vehicle's dynamic, the road geometry, the dimension of the vehicle or the occupancy of objects in the environment.
Model Predictive Control (MPC) for example sets the current control by anticipating future events using a model of the system dynamics. It is originally used as a control method but have been extended in the literature for decision-making. In~\cite{nilsson2013strategic}, it is presented a decision and control algorithm for lane change and overtaking maneuvers. The problem of deriving decisions regarding appropriate driving maneuvers i.e., selection of desired lane and velocity profile, on two-lane, one-way roads, is considered as a Mixed Logical Dynamical (MLD) system to be solved through MPC using mixed integer program formulation. The predictive controller allows full control of acceleration/deceleration as well as providing a decision variable regarding preferred lane at each time instant.

Today's research in decision-making focuses mostly on finding global and robust solutions. Solutions that generalize to all situations while taking into account uncertainty, unpredictable situations while guaranteeing always the AV safety.

\subsection{Decision-making based AI}\label{sec:ia_dec}
In contrast with the methods shown in the previous section, where decision are either encoded by hand or directly nested in the planning, in this section will be discussed methods based on either a probabilistic definition or learning-based approach.
\subsubsection{Probabilistic approaches}
Among the main used probabilistic framework for decision-making in the literature let us mention:
\begin{itemize}
\item Markov Decision Processes (MDP): Partially Observable MDP (POMDP), Mixed Observability MDP (MOMDP), Hidden Markov Model (HMM) and semi-MDP (sMDP)~\cite{brechtel2011probabilistic,gindele2010probabilistic,Ulbrich2013,hubmann2017decision}.
\item Bayesian Networks (BN): Two types are used the Dynamic Bayesian Network (DBN) and the Decision Network (DN) (also called Influence Diagrams)~\cite{schubert2012evaluating,schulz2018multiple,forbes1995batmobile,iberraken2018safe}.
\end{itemize}

\paragraph{Methods based on Markov Decision Processes (MDP)}
MDP is a discrete-time stochastic state transition system~\cite{bellman1957markovian,howard1960dynamic}.

It is used to model the sequential decision process of an agent acting in a dynamic environment with uncertain dynamics. In an MDP, the agent interacts with its environment - while taking the available information of the state of the world- by taking actions at discrete time steps. Upon taking such actions, the state of the world changes and make a transition and the agent receives a reward signal.
 
The goal of such a problem is to find an optimal policy (sequence of actions) $\pi^*$ that maximizes the expected reward over the time horizon. The commonly applied approach to find an optimal policy is value iteration~\cite{bellman1957markovian}.

Mouhagir et al., in~\cite{Mouhagir16} proposed a method based on an MDP like model for trajectory planning with clothoid tentacles. The idea is to generate realistic trajectories with tentacles method and select the best tentacle regarding the MDP process. However, the simulations was held with only static obstacles with no uncertainty consideration.
In~\cite{brechtel2011probabilistic} a theoretical approach is proposed for combining continuous world prediction by a DBN and discrete world semi-MDP planning. A semi-MDP allows actions that take varying amount of times to complete and is very useful in this work where the goal is to plan sequences of lane change maneuvers. The transition probabilities of the semi MDP are modeled using the DBN proposed by Gindele et al.,~\cite{gindele2010probabilistic} that accounts for the interactions between vehicles. The policy is then recalculated each time a new traffic situation is encountered, which still raises the question of real-time applicability.\\
POMDP~\cite{drake1962observation}, on the other hand, is an extension of an MDP to account for partial observability of states in a system. POMDP helps to introduce the idea of a belief $bel(x_t)$ of being in a state $x_t$ at time $t$.

The objective of solving a POMDP is to find an optimal policy $\pi^*$ which maximizes the expectation for the reward sum over the future time steps. However, despite the close relationship between POMDP and MDP, solving a POMDP is considerably more difficult than solving the corresponding MDP as the POMDPs computational complexity grows exponentially with the planning horizon. To overcome this issue, approximate solution methods have been proposed. Some of the solutions focus on solving offline POMDP models which means that the focus is not to calculate the best possible action for the current belief state but rather for every imaginable belief state. This restricts their applicability to only small problem domain. In contrast, online approaches~\cite{Ulbrich2013, hubmann2017decision} allow a calculation of a good policy at the current belief state of the agent. In what follows, some works using these methodologies for decision-making are presented.
In~\cite{Ulbrich2013}, online POMDP is used for decision-making for performing lane changes while driving fully automated in urban environments. The online POMDP is applied to accommodate inevitable sensor noise to be faced in urban traffic scenarios. An ingenious way is proposed to keep the complexity of the POMDP low enough for real-time decision-making while driving through a two steps algorithm. The first step is through signal-processing networks that assesses the situation, whether a lane change is feasible or not, and whether a lane change is advantageous or not. The outputs of this signal processing networks are submitted in the second step to the POMDP decision-making algorithm.

Brechtel et al., in~\cite{brechtel2014probabilistic} in the other side presented a generic approach for decision-making under uncertainty through a continuous POMDP that can be optimized for different scenarios. The cornerstone of this work consists in considering uncertainties and finding the suited space representation for driving as the prevailing challenges for automatic decision-making. For this reason, the proposed POMDP automatically learns a suited representation depending on the specific given problem which is directly based in this paper on the pose and velocities of the involved road users.
We can see through the above mentioned works that the use of POMPD differs in the way of solving them in terms of the used methodology, the complexity of the design and the driving environment. \\
A Hidden Markov Model (HMM) is a temporal probabilistic model in which the state of the process is described by a single discrete random variable. The possible values of the variable are the probable states of the world. 
The work given in~\cite{laugier2011probabilistic} proposed to use HMMs for the behavioral model with the aim of estimating the probability for a vehicle to perform one of its feasible behaviors. The behavioral model comprises two hierarchical layers, and each layer consists of one or more HMMs. The upper layer is a single HMM where its hidden states represent high-level behaviors, such as: move straight, turn left, turn right, or overtake. For each behavior in the HMM of the upper layer, there is a corresponding HMM in the lower layer which represents the sequence of state transitions of the behavior, for example \textit{Overtake} has 4 hidden states: lane change, accelerate (while overtaking a car), lane change to return to the original lane, resume a cruise speed. 

\paragraph{Methods based on Bayesian Networks (BN)}
The second well known family of probabilistic decision-making is based on BN. BNs for decision-making is used through two known extensions. The first one is Decision Network (DN) or Influence Diagram and the second is DBNs.
Bayesian Networks in summary are Directed Acyclic Graphs (DAG) in which each node corresponds to random variables connected by directed links called arcs. For every variable $X_i$ in the graph, with parents $X_j$ Bayes theorem is applied to quantify the effect of the parents on the node and deduce the conditional probability distribution.     

The aforementioned conditional probabilities are summarized in a conditional probability table (CPT). The topology of the network specifies the conditional independence relationships between variables which makes Bayesian Networks by definition computationally tractable for reasonably small networks. This is one of the main advantages of Bayesian Networks. 
BNs are used for probabilistic reasoning which is a method of representation of knowledge where the concept of probability is applied to indicate the uncertainty in knowledge. It is largely used in many industries and according to~\cite{russell2016artificial} BNs are a leading paradigm in AI research on uncertain reasoning and expert systems. It allows for learning from experience, and it combines the best of classical AI and neural networks. \\
Decision Networks (DNs or Influence diagram)~\cite{russell2016artificial} combine BNs with additional node types for actions and utilities. DNs allow us to support probabilistic reasoning, decision-making under uncertainty for a given system and yield the capacity to incorporate multiple decision criteria.
In the field of decision-making for autonomous driving many use DNs.
When modeling the DNs, most of the work chose to design the topology of the BNs with two main levels: the situation assessment level to infer the current situation state based on the risk assessment and the decision-making strategy to deduce the maneuvering decisions.
Using this formalism, Schubert in~\cite{schubert2012evaluating} uses DNs for lane change decision-making. The paper presents a system that can perceive the vehicle’s environment, assess the traffic situation, and gives recommendations about lane-change maneuvers to the driver. The situation assessment is done in the upper layer while using a well-known threat measure the Deceleration To Safety Time (DST) as a threat measure to assess the danger of the navigation lanes status. The lower layer is dedicated to decision-making with one decision node for lateral maneuvers. The utility value is assigned manually for each combination of traffic situation and the maneuver decision.\\
A DBN on the other hand is a Bayesian network that represents a temporal probability model. The system is modeled as series of snapshots or time slices each of which contains a set of random variables. DBNs generalize HMMS by allowing  the state space to be represented in factored form, instead of a single discrete random variable.
DBNs are extensively used for maneuver intention, trajectory prediction and modeling the interaction between traffic participants (cf. Table~\ref{tab:2}) which makes them very suitable for decision-making.
One of the pioneering contribution and a must cited example is the work of Forbes et al.,~\cite{forbes1995batmobile} with the BATmobile.
The authors proposed to use a Dynamic Probabilistic Network (DPN). The used DPN resembles the definition of a DBN and contains nodes for sensor observations as well as nodes for predicting driver intentions, such as whether the driver intends to make a lane change or to slow down. The DPN is used as the basis for three separate decision-making approaches: dynamic decision networks which is the DPN extended with actions node and utility function for each time slice, hand-coded policy representations through a decision tree and supervised learning and reinforcement learning methods for solving the full POMDP.
Even though Forbes et al., deduced that avoiding manual programming and considering partial observability improves the results, their proposed solution were ahead of their time given the present state of technology.
Schulz et al.,~\cite{schulz2018multiple} proposed a decision-making framework, which explicitly considers the intentions of drivers and the inter-dependencies between their future  behaviors. The decision-making process of the agent is divided into three hierarchical layers: which route it is going to take (route intention $R_t$),  whether it is going to pass a conflict area at an intersection before or after another agent  (maneuver intention $M_t$), and what continuous action $A$ is going to executed.
The proposed DBN have two consecutive time slices and the inter-dependencies between vehicles $V_i$.
The used inference algorithm is a Multiple Model Unscented Kalman Filter (MM-UKF). Each UKF represents the complete state space, i.e., kinematic state, route, maneuver, and action of all agents in the scene.

\subsubsection{Learning-based approaches}\label{sec:learn_dec}
In this section we will focus only on the methods developed specially for decision-making using learning approaches as end-to-end techniques have been detailed in section~\ref{sec:endtoend}. Three key paradigms in machine learning exist: Supervised, Unsupervised and Reinforcement. Supervised learning makes decision based on the output labels provided in training. Two commonly used examples of supervised learning for decision-making are Decision Trees and Random Forest. Unsupervised learning is based on unlabeled data. It aims to find an accurate representation of the unlabeled information. Clustering is a type of unsupervised learning that gathers samples of similar characteristics. The third machine learning paradigm is Reinforcement Learning (RL), that takes its root from Markov decision process (MDP). In RL the agent interacts with the environment to learn how to behave without having any prior knowledge by learning to maximize a numerically defined reward (or to minimize a penalty). 
In this regard, Ngai et al., in~\cite{ngai2011multiple} proposed a reinforcement learning algorithm of lane change decisions for highway driving. By making use of Q-learning for determining action decisions, seven different goals are considered among which: lane changing, collision avoidance, and lane following, as the author believes that by expliciting the goals, the problem can be better solved. 
The Inverse Reinforcement Learning (IRL) model~\cite{kuderer2015learning} have also been used to get the individual driving style of traffic participants and plane the safest trajectory. In a similar fashion, Q-learning algorithm have been applied for lane change scenarios~\cite{you2018highway}. \\
Deep Learning (DL) is closely related to the above three paradigms of ML and is used to extract higher-level features from data. DL are inspired by the multi-layered structure of human neural system and recurrent neural networks and convolutional neural networks are examples of known deep learning architectures. 
Human-like decision-making based on DL have been extremely used in the field of ITS~\cite{haydari2020deep,grigorescu2019survey,tai2016deep,arulkumaran2017brief} and robotics in general~\cite{machkour2022classical}. Contrary to the other methods for decision-making, these methods recognize human personality and social intelligence and does not fully focus on the ``correctness''~\cite{sheridan2016human} as they learn from real driving scenarios. According to Waymo, accidents that occurred with their AVs can also be used as a valuable experience for the self-driving system~\cite{yuan2016suppose}. On this basis, several neural network have been proposed for the decision-making strategies. In~\cite{li2018humanlike} a decision-making system is presented, the main novelty lies in the human-like thinking ability integrated in the neural network.
RL combined with deep learning, named deep RL, is according to~\cite{haydari2020deep} currently accepted as the state-of-the art learning framework in control systems. While RL can solve complex control problems, deep learning helps to approximate highly nonlinear functions from complex dataset.\\
The mentioned deep learning methods do offer great advantages in terms of flexibility and scope of utilization. However, one of the main drawback is that no one can analytically ensure that the corresponding output for these systems will tend towards always an acceptable safe solution. For this reason, some works choose to combine deep learning approaches with safety verification methods to guarantee safety of the decisions. Mirchevska et al., in~\cite{mirchevska2018high} present a reinforcement learning-based approach, that is combined with formal safety verification to ensure that only safe actions are chosen at any time. The deep reinforcement learning agent learn to drive as close as possible to a desired velocity by executing reasonable lane changes on highways.
However, this doesn't resolve another drawback of deep neural networks concerning the computational complexity needed in the learning phase and the fact that these methods do not have any analytical formulation, which leads to non-provable outputs.

\subsection{Guarantee of safety of AVs}\label{sec:safety_ver}
Despite several years of developments of decision-making strategy for autonomous vehicles (AV) and the rich literature in this domain, there is unfortunately not yet a fully generic solution that deals with all kinds of scenarios. For this reason, recent advances in AVs raised all the importance of ensuring the complete safety of AV maneuvers even in highly dynamic and uncertain environments/situations. This objective becomes even more challenging due to the uniqueness of every traffic situation/condition. Indeed, the lack of safety guarantees proves, which is one of the key challenges to be addressed, limit drastically the ambition to introduce more broadly AVs in our roads, and restrict the use of AVs to very limited use cases.\\
This section provides a discussion on the methods used to guarantee the safety of AVs among which it is investigated: safety verification techniques, evasive maneuvering in emergency situations and the standardization/generalization of safety frameworks. 

\subsubsection{Safety verification}
The common task for AVs after the decision-making is to determine a nominal trajectory to perform lane changes and other maneuvers, taking into consideration any constraints or traffic condition that are known at the time of planning. The procedure must therefore be aborted automatically in case of any unexpected approaching objects, such as other objects and road users, entering the planned course of the vehicle. The vehicle must then be able to replan by determining an alternate route, i.e., the emergency trajectory, which the car will pursue instantly to avert an accident and guarantee safety all the time.\\
Extensive testing to simulate all possible behaviors of other traffic participants is a time-consuming task. Indeed, considering the uniqueness of each traffic situation, the task of modeling every situation is nearly impossible. In addition, it can only prove that a system is unsafe, but is not able to propose an alternative.
Classical safety verification techniques perform the safety verification offline before the vehicle is deployed and can only investigate a certain class of situations, as it is done for the verification of an automatic cruise controller in~\cite{stursberg2004verification}. These techniques are  usually called \textit{scenario-based verification}.\\
Since every traffic situation is unique, it is necessary that the decided/planned maneuvers be always verified during navigation of the vehicle. This has been called in the literature online safety verification~\cite{althoff2014online} or formal verification and answers to this challenge. It has been used in many works of the literature~\cite{mitsch2012towards,althoff2014online}. 
In~\cite{althoff2014online} reachability analysis is used as a safety verification method. The verification is carried out by estimating all potential occupancies set of the automated vehicle and other traffic participants. To capture all potential future possibilities, reachability analysis is applied to account for all potential behaviors of mathematical models including uncertain inputs (e.g., sensor noise, disturbances) and partially unknown initial states. Possible future collisions are identified when comparing the intersection of the obtained sets. However, if the trajectory is regarded as unsafe no alternative is proposed to avoid the collision.\\
Pek et al., in~\cite{pek2017verifying} presented an approach for verifying the safety of lane change maneuvers of AVs by incorporating formalized traffic rules. The assessment is based on safe distances, allowing the ego vehicle to drive safely for an infinite time horizon which allows the AV to verify its decision to change lanes and recover if the lane change becomes unsafe during the maneuver.\\
Safety Verification of Deep Neural Networks on the other hand, have been little studied and few works exist in this topic. Huang, et al.,~\cite{huang2017safety} proposed a framework for automated verification for the safety of classification decisions, which is based on search for an adversarial misclassification within a given region. The key distinctive features of this framework compared to existing work is the guarantee that a misclassification is found if it exists.
In a global manner, deep learning techniques have become increasingly popular in the domain of decision-making and autonomous navigation, however, still remains questions about their ability to guarantee safety since their output responses are not well known, particularly outside the training data scope. Authors in~\cite{seshia2016towards} identified some challenges about the safety verification process in artificial intelligence systems and they mainly consist in modeling issues of the environment or the system as formal verification critically relies on having a precise, mathematical statement of what the system is supposed to do.

\subsubsection{Evasive maneuvering in emergency situations}\label{sec:litter_evasiv}
Because maneuvers are verified online while using safety verification techniques, the ability of the system to re-plan and evade a dangerous situation becomes possible. Emergency scenarios may necessitate maneuvering up to the vehicle’s handling limits in order to avoid collisions~\cite{funke2016collision}.
The common used methods and the one from very early work related to emergency situations is to simultaneously plan a nominal and an emergency trajectory in order to guarantee the safety of the vehicle controller. With the help of this planning process the vehicle controller is able to provide an emergency trajectory before and during the performance of a lane change or any other maneuver. Several literature research tackle this problematic~\cite{ hirsch2005optimization, magdici2016fail}. However, generating an emergency maneuver for each time step is computationally expensive and often not needed. \\
Presented in~\cite{bae2014decision} a Nonlinear Model Predictive Control (NMPC) dedicated for emergency collision avoidance in complex situations between a driver's vehicle and neighboring vehicles. NMPC is used to predict the trajectories of all vehicles, and if a collision is detected in the predicted trajectories, the driver's vehicle attempts to avoid the collision through a left lane change, right lane change, or braking maneuver. 
An even detailed procedure for evasive action handling is proposed in~\cite{pek2018computationally}. Pek et al., developed a fail-safe trajectory planner for self-driving vehicles. This trajectories are computed in real-time in continuous space by making use of convex optimization techniques. This approach simultaneously favors jerk minimization through the defined motion models and their corresponding constraints and incorporates the safety verification in the planner in such a way that trajectories are always verified as safe. However, this approach lacks real experimentation to prove its efficiency.
Another interesting approach is proposed in~\cite{Iberraken2019reliable} where a Sequential Decision Networks for Maneuver Selection and Verification (SDN-MSV) that utilizes multiple complementary threat measures to propose discrete actions that allow to: derive appropriate maneuvers in a given traffic situation, provide a safety retrospection and verification over the current maneuver risk and outputs, if advised, appropriate evasive action according to the environment dynamic, in order to face any sudden hazardous and risky situation. However, the evasive decision has been applied to the system with a constant velocity configuration while having an already defined fixed path to follow, which limits the flexibility of the evasion.\\
A step forward into emergency situation management, is considering cases where the collision is unavoidable. 
Fraichard et al., in~\cite{fraichard2004inevitable} have worked on the subject by proposing a concept called Inevitable Collision State (ICS). An ICS is characterized as a state for which, no matter what the future trajectory followed by the system is, a collision with an obstacle eventually occurs. It takes into consideration both the dynamics of the system and the obstacles. The concept is useful both for navigation and motion planning purposes as for its own safety, a robotic system should never find itself in an inevitable collision state, however, determining ICSs is computationally intense and suffer from uncertain future motion of obstacles.
To palliate to this issue, authors in~\cite{pek2018efficient} introduce invariably safe sets which are regions that allow vehicles to remain safe for an infinite time horizon. A tight under-approximation of the proposed sets is obtained in real-time with respect to the number of traffic participants while maintaining formal safety guarantees. Moreover, these sets have been used to determine the existence of feasible evasive maneuvers and the criticality of scenarios by computing the time-to-react metric. The drawback around set-based methods is that they tend to be overly conservative. Thus, in dense traffic and a highly uncertain environment, these methods are likely to yield no suitable solution.

\subsubsection{Standardization and generalization of safety frameworks}
In order to be compared, decision-making framework must be tested on the same data set. To the best of our knowledge, no publicly available data set exists with real nominal/emergency situations on which decision-making algorithm has been tested and published. Simulation could be used as a mean to generate collision data which could be used by the ITS community to compare algorithms. In addition, in this matter no standardize performance indicator exists to be able to compare algorithms between them.  \\
An interesting resolution of the current years is the willingness of IEEE to start a range of recently recognized AV working groups. 
Under the reference IEEE P2846~\footnote{https://sagroups.ieee.org/2846}, the IEEE standardization organization has started working on the definition of ``A Formal Model for Safety Considerations in Automated Vehicle Decision-Making''. The idea for industries and governments is to be able to eventually align with a common definition of what it means for an AV to make decisions that balances between safety and practicality.
According to the IEEE, this approach is justified by the fact that the decision-making capacity of an on-board computer, with its set of artificial intelligence algorithms, is generally hidden from observation and constitutes a sort of ``black box''. This makes an objective comparison of the safety offered by different AVs almost impossible. As some experts have already pointed out, the IEEE emphasizes that statistical evidence - such as the number of kilometers traveled, the frequency of human intervention or the hours of simulation - cannot capture all situations, especially those that the AV has never seen before. In concrete terms, the future IEEE P2846 standard aims to define a formal mathematical model based on rules related to vehicle decision-making using mathematical algorithms and discrete logic. The model will apply to the planning and decision-making functions of an AV from levels 3 to 5 (according to the SAE standard grading for vehicle automation)~\cite{sae2014taxonomy}.
The model will be formally verifiable, via mathematical proof, will be technology-neutral and will be parameterizable to ensure the necessary customization at the level of individual jurisdictions. The standard will apply to specified scenarios and driving cases that do not eliminate all hazards, but that strike a balance between safety on the one hand and reasonable feasibility of application on the other. \\
A highly publicized contribution by Mobileye~\cite{shalev2017formal} proposed a standardization of safety assurance and a formal model of safety by answering two main challenges: lack of safety guarantees, and lack of scalability. The lack of safety guarantees relies on answering the question what are the minimal requirements that every self-driving car must satisfy? and how can these requirements be verified? The second area of risk scalability concerns engineering solutions that result in huge costs will not be scalable to millions of cars, which will push interest in this field into a niche academic corner, and drive the entire field into what the authors called a “winter of autonomous driving”. The combined issues leads the authors to propose a number of properties that answers to these challenges gathered in a framework called the Responsibility-Sensitive Safety (RSS) and that covers all the important ingredients of an AV: sense, plan and act. The RSS represent a rigorous mathematical model formalizing an interpretation of the law which is applicable to self-driving cars thus guarantees that from a planning perspective there will be no accidents which are caused by the AV. In addition, the framework has been designed such that it is not overly-defensive concerning the driving policy and efficiently verifiable in the sense that it can be proved that the self-driving car implements correctly the interpretation of the law. Jack Weast, Senior Principal Engineer at Intel has been appointed by the IEEE P2846 to lead the working group responsible for building the standard. To get started, Intel will contribute with its Responsibility-Sensitive Safety (RSS) environment. 
\subsection{Discussion}
This section discussed the state of the art algorithms in the domain of decision-making of AVs. Table~\ref{tab:decision} summarizes the main characteristics of the aforementioned decision-making algorithms.\\
\begin{table*}[th!]
\begin{center}
 \resizebox{\textwidth}{!}{%
\begin{tabular}{p{3.5cm}|p{3cm}|p{2cm}|p{2.5cm}|p{2.5cm}|p{3cm}|p{3cm}|p{2.5cm}|p{3cm}}
\toprule
\textbf{Approach}       &\textbf{Reference} &\textbf{Prediction}     &\textbf{Uncertainty Handling} & \textbf{Offline/Online}  &\textbf{Space} & \textbf{Computational complexity} &\textbf{Generalization} &\textbf{Online Safety Verification} \\ \midrule

\multirow{1}{*}{\textit{\textbf{Rule-based DM}}} &  FSM~\cite{kress2008automatically} &  X &  X  &  Offline   &   Discrete   & Low &   - -  & X \\   \midrule  
\multirow{5}{=}{\textit{\textbf{Optimization-based DM}}} & Trajectory Planning~\cite{ziegler2014making}    &  \checkmark &  Uncertainty in the measurements  &  Online   &  Continuous   & Manageable &  +  & X  \\ \cline{2-9}
& Fail-safe Trajectory Planning~\cite{pek2018computationally}    &  $\checkmark^b$ &  Uncertainty in the measurements  &  Online   &  Continuous   & Manageable &  + & \checkmark \\ 
                                      \midrule

\multirow{10}{=}{\textit{\textbf{Probabilistic approaches}}}   & Decision Networks~\cite{schubert2012evaluating} &  \checkmark &  Uncertainty in the states &  Offline  &   The states are either discrete or continuous, it uses a utility value for decision   & Low &  + &  (\checkmark) \\ \cline{2-9}
                          &  POMDP~\cite{brechtel2014probabilistic,Ulbrich2013}   & $\checkmark^{a,b} $&  Uncertainty in the states and measurements &  Online  &   Continuous & High & + + & X \\     \cline{2-9} 
                         & DBN~\cite{schulz2018multiple} &    $\checkmark^{a,b}$   &  Uncertainty in the states and measurements &   Online  &  Continuous & High & + +  & (\checkmark)  \\ 
                        \midrule
\multirow{1}{=}{\textit{\textbf{Learning-based approaches}}} & Reinforcement Learning~\cite{ngai2011multiple,you2018highway} & (\checkmark) & (\checkmark)  &  Offline  & Discrete & High &  -  & X\\

\bottomrule

\end{tabular}}
\caption{Comparison of Decision-Making (DM) approaches for AVs. X means that the feature is not supported. ($\checkmark$) means that the feature is not supported in the original work but can be integrated. $\checkmark^a$ means the prediction considers interaction between traffic participants. $\checkmark^b$ means that long time horizon prediction is considered. Offline/Online means whether the system find the best possible maneuver to be executed in the current situation or during an offline training phase. }\label{tab:decision} 
\end{center}
\end{table*}
The most important aspects in a decision-making framework is its ability to solve any situation, consider uncertainty and unexpected situation while finding the right balance between accuracy and computational expenses.\\
The first distinction between them lies in the uncertainty handling. The most complete methods take into account the uncertainty in the states and measurements of the traffic environment. The second distinction relies in their real-time execution ability (learns or reason during an offline training phase and execute the maneuver online in the current driving situation). Due to the unlimited number of traffic situations (especially that training data for emergency situation is scarce) the offline training can be unsatisfactory and insufficient. However, if this kind of data become available, learning-based approaches can become the best solution to autonomous driving. Another important aspect is online safety verification and this is where difference can be made compared to other approaches. Probabilistic approaches stand up well regarding the defined characteristics as it have the potential to consider the nature of the stochastic dynamics of a traffic environment, be able to account for uncertainties through well-known probabilistic algorithms and consider for present and future interactions between participants. The model-based and probabilistic approaches allows also to easily include an online verification mechanism in the decision-making of self driving cars due their analytical definition. This latter is mandatory since every traffic situation is almost unique and a quick response is needed to deal with any emergency situation.

\section{Conclusion} \label{sec:conclusion}
This paper has surveyed research on autonomous vehicles while focusing on the important topic of safety guarantee of AVs. Scrutinizing through the literature, it is presented a detailed review of relevant methods and concepts defining an overall control architecture for AVs, with an emphasis on the safety assessment and decision-making systems composing these architectures. Moreover, through this reviewing process, it is highlighted research that uses either model-based methods or AI-based approaches. This is performed also while emphasizing the strengths and weaknesses of each methodology and investigating the research that proposes a comprehensive multi-modal design that combines model-based and AI approaches. With these investigations, it was shown one of the promising ways to reach the mentioned requirements and characteristics that can be obtained by a smart combination of model-based methods and AI-based approaches when applied in a coherent, complementary, and synergistic manner. 
This paper ends with a discussion of the methods used to guarantee the safety of AVs among which it is investigated: safety verification techniques and the standardization and generalization of safety frameworks. These subjects remain regardless of the used method a challenge in the AV domain as autonomous vehicles are probably the most advanced intelligent systems under development so far in the world.

\section*{Declarations}

\subsection*{Funding}
This work was sponsored by a public grant overseen by the French National Research Agency as part of the “Investissements d’Avenir” through the IMobS3 Laboratory of Excellence (ANR-10-LABX-0016) and the IDEX-ISITE initiative CAP 20-25 (ANR-16-IDEX-0001).

\subsection*{Code or data availability (software application or custom code)}
Data sharing not applicable to this article as no datasets were generated or analysed during the current study.
\subsection*{Competing Interests}
The authors have no relevant financial or non-financial interests to disclose.
\subsection*{Author Contributions}
All authors including Dimia Iberraken and Lounis Adouane contributed to this study. The first draft of this manuscript was written by Dimia Iberraken. Both authors were active in reviewing the submitted manuscript and approved it.

\subsection*{ Ethics approval}
This research work did not involve human participants or animals. Hence, ethics approval, Consent to participate and Consent for publication are not applicable.
\subsection*{ Consent to participate}
Not applicable
\subsection*{ Consent for publication}
Not applicable



\bibliography{BiblioThesis}


\end{document}